\RequirePackage{fix-cm}
\documentclass[11pt,draftcls,onecolumn]{IEEEtran}

\ifCLASSINFOpdf
  \usepackage[pdftex]{graphicx}
\else
\fi
\usepackage{url}
\usepackage[
CJKbookmarks=true, bookmarksnumbered=true, bookmarksopen=true,
colorlinks=true, 
pdfborder=001, 
citecolor=blue, linkcolor=green, anchorcolor=red, urlcolor=blue
]{hyperref}

\usepackage{amssymb}
\usepackage{algorithm}
\usepackage[cmex10]{amsmath}

\usepackage{array}

\usepackage{bbding}
\usepackage[cmex10]{amsmath}
\usepackage{url}
\usepackage{color}
\usepackage{hyphenat}
\usepackage{multirow}

\usepackage[tight,footnotesize]{subfigure}

\begin{document}
%
\title{Random Sampling for Fast Face Sketch Synthesis}
%
%
%

\author{Nannan~Wang,
        and Xinbo~Gao,
       and Jie~Li
\thanks{
N. Wang is with the State Key Laboratory of Integrated Services Networks, School of Telecommunications Engineering, Xidian University, Xi'an 710071, Shaanxi, P. R. China (e-mail: nnwang@xidian.edu.cn).

X. Gao and J. Li are with the State Key Laboratory of Integrated Services Networks, School of Electronic Engineering, Xidian University, Xi'an 710071, Shaanxi, P. R. China (e-mail: xbgao@mail.xidian.edu.cn;leejie@mail.xidian.edu.cn).
}
}

\maketitle

\begin{abstract}
Exemplar-based face sketch synthesis plays an important role in both digital entertainment and law enforcement. It generally consists of two parts: neighbor selection and reconstruction weight representation. The most time-consuming or main computation complexity for exemplar-based face sketch synthesis methods lies in the neighbor selection process. State-of-the-art face sketch synthesis methods perform neighbor selection online in a data-driven manner by $K$ nearest neighbor ($K$-NN) searching. Actually, the online search increases the time consuming for synthesis. Moreover, since these methods need to traverse the whole training dataset for neighbor selection, the computational complexity increases with the scale of the training database and hence these methods have limited scalability. In this paper, we proposed a simple but effective offline random sampling in place of online $K$-NN search to improve the synthesis efficiency. Extensive experiments on public face sketch databases demonstrate the superiority of the proposed method in comparison to state-of-the-art methods, in terms of both synthesis quality and time consumption. The proposed method could be extended to other heterogeneous face image transformation problems such as face hallucination. We release the source codes of our proposed methods and the evaluation metrics for future study online: \url{http://www.ihitworld.com/RSLCR.html}.
\end{abstract}


\IEEEpeerreviewmaketitle

\section{Introduction}
\label{sec:Introduction}

Face sketch synthesis mainly refers to generating a sketch given one input photo and some face sketch-photo pairs as the training dataset. It has achieved wide applications in both digital entertainment and law enforcement \cite{Ref0}. For example, since limited information about the suspect is available due to the low quality of surveillance videos or even no video/image clues, a sketch drawn by the artist is usually taken as the substitute for suspect identification. Then, face sketch synthesis bridges the great texture discrepancy between face photos and sketches.

Exemplar-based face sketch synthesis generally proceeds in two steps: neighbor selection and reconstruction weight representation. Given an input test photo, it is divided into some patches with even size and adjacent patches have some overlap to guarantee the compatibility. Then for each test patch, some number (\textit{e.g.} $K$) of nearest photo patches are selected from the training photos. Sketch patches corresponding to these nearest photo patches are taken as the candidates for sketch patch synthesis. The prevalent way to represent the target sketch patch is the linear combination of selected candidate sketch patches. The linear combination coefficients are usually calculated under the assumption that a photo patch and its corresponding sketch patch share similar geometric manifold structure, \textit{i.e.} if two photo patches are similar, then their sketch patch counterparts are also similar.

Exemplar-based face sketch synthesis started from the work of Eigen-transformation of Tang and Wang \cite{Ref3,Ref4}. In their work, there is no special neighbor selection process but all training images are utilized. The linear combination coefficients are learned by projecting the input photo onto the training photos through principal component analysis.

Considering that only learning one holistic reconstruction model is difficult to represent the nonlinear mapping between face photos and sketches, Liu \textit{et al.} \cite{Ref6} proposed to estimate the holistic nonlinear mapping relationship with many piece-wise linear mappings, which are generally followed in subsequent methods. This method works on the image patch level. $K$ nearest photo patches are searched from the training set in terms of Euclidean distance. Then the reconstruction weight is calculated in the spirit of locally linear embedding \cite{Ref7}:
\begin{equation}
\label{eq:1}
\min_{\mathbf{w}}\|\mathbf{x}-\mathbf{Xw}\|_2^2, s.t. \sum_{i=1}^K{w_i}=1
\end{equation}
where $\mathbf{w}=(w_1,w_2,\cdots,w_K)^T$ is the representation weight vector, $\mathbf{x}$ is the test photo patch in the form of column vector and $\mathbf{X}$ is the matrix of column-concatenation of $K$ selected training photo patches. The target sketch patch corresponding to the test photo patch is reconstructed from the linear combination of $K$ training sketch patches weighted by $\mathbf{w}$. Song \textit{et al.} \cite{Ref1} casted the face sketch synthesis problem into a spatial sketch denoising (SSD) problem and calculated the reconstruction weight through conjugate gradient solver. Gao \textit{et al.} \cite{Ref8} proposed to adaptively determine the number of nearest neighbors by sparse representation \cite{Ref9} rather than the fixed number (\textit{e.g.} $K$) of nearest neighbors. Instead of using sparse representation for neighbor selection, some dictionaries are learned through sparse coding and sparse representation to substitute the role of nearest neighbors in the work \cite{Ref10}.

Wang \textit{et al.} \cite{Ref13} employed Markov random field (MRF) to model the dependency from two aspects: the dependency between test photo patches and nearest photo patches and the dependency between adjacent synthesized sketch patches which are neglected in the above methods. In their method, $K$ nearest photo patches and their corresponding sketch patches are selected from the training dataset. Only one single nearest sketch patch is finally selected through MRF networks which is taken as the target synthesized sketch patch. In other words, the weight reconstruction representation for this method can be seemed as finding one most appropriate sketch patch and its weight is set to 1.

Zhou \textit{et al.} \cite{Ref15} proposed to introduce the linear combination into the MRF model (namely Markov weight field, MWF) to overcome the face deformation problem due to single sketch patch search strategy in \cite{Ref13}. The difference between the MWF method and the LLE method \cite{Ref6} is the consideration of the dependency between adjacent synthesized sketch patches as follows:
\begin{equation}
\label{eq:2}
\begin{aligned}
&\min_{\mathbf{w}}\|\mathbf{x}-\mathbf{Xw}\|_2^2+\lambda\|\mathbf{O}^i\mathbf{w}-\mathbf{O}^{(i)}\mathbf{w}^i\|,\\
&s.t. \sum_{l=1}^K{w_l=1}, w_l\geq0,
\end{aligned}
\end{equation}
where the second term $\sum_{i=1}^4\|\mathbf{O}^i\mathbf{w}-\mathbf{O}^{(i)}\mathbf{w}^i\|$ represents the dependency constraint between the synthesized sketch corresponding to the test photo patch $\mathbf{x}$ and its four adjacent synthesized sketch patches. Here the constraint is modeled by the distance of pixel intensity vector extracted from the overlapping area between adjacent sketch patches. In equation (\ref{eq:2}), the column vector $\mathbf{w}^i$ is the reconstruction weight corresponding to the $i$-th adjacent sketch, $\mathbf{O}^{(i)}$ denotes the pixel intensity vector extracted from the overlapping area of adjacent sketch patch $i$, and $\mathbf{O}^i$ denotes the pixel intensity vector extracted from the overlapping area of current target sketch patch. Wang \textit{et al.} \cite{Ref16} further developed the MWF model from the perspective of transductive learning. Peng \textit{et al.} \cite{Ref17} extended the MWF model to a multi-view version which improves the robustness against the cluttered background and lighting variations. Unlike the even patch employed in aforementioned methods, super-pixel segmentation of image patches is employed and the reconstruction representation model as in equation (\ref{eq:2}) is adopted in the work \cite{Ref29}.

All aforementioned methods perform $K$ nearest neighbor ($K$-NN) selection online, which heavily increases the time consuming for test. Moreover, with the increase of the scale of the database, the computation complexity would also increase linearly. In addition, the reconstruction weight representation model either in (\ref{eq:1}) or in (\ref{eq:2}) consider that all selected nearest neighbors contribute equally to the reconstruction weight computation process, while the distinct distance between these neighbors and the test patch are neglected.

\begin{figure*}
\centering
\includegraphics[width=1\columnwidth]{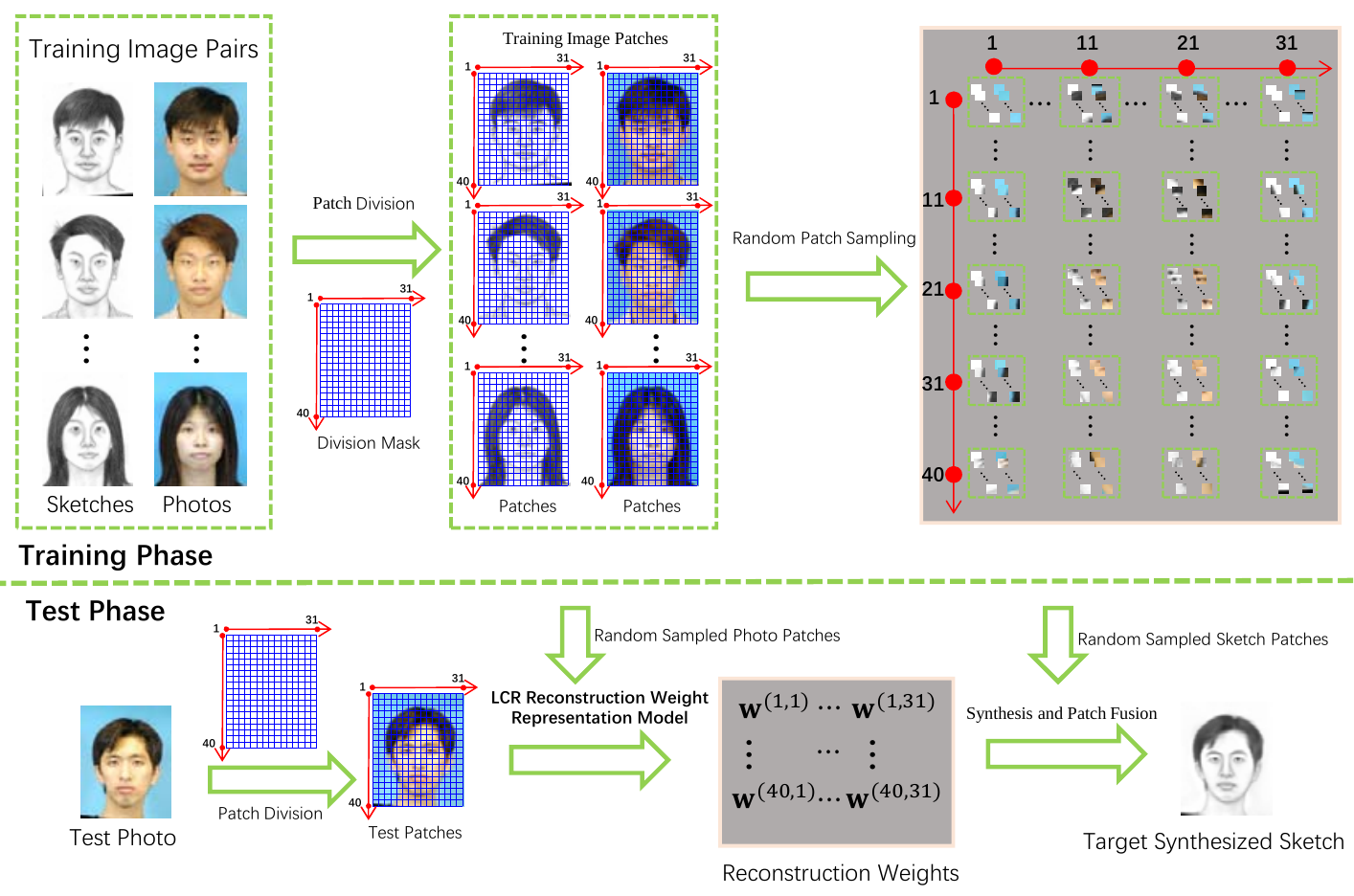}
\caption{A graphical illustration of the proposed RSLCR framework.}
\label{fig:1}
\end{figure*}

In this paper, instead of online searching neighbors, we randomly sample some patches offline and then these patches are used to reconstruction the target sketch patch. This random sampling strategy greatly speeds up the synthesis process, which is much faster than $K$-NN based methods (\textit{e.g.} the LLE method \cite{Ref6}) under the same experimental settings.  In addition, state-of-the-art methods consider that all selected neighbors contribute equally to the reconstruction weight computation process while the distinct similarity between the test patch and these neighbors are neglected. Since these random sampled patches have distinct similarities with the test photo patch, we impose the locality constraint \cite{Ref30} to regularize their corresponding reconstruction weights. The locality constraint would restrain the contribution of patches which distribute far from the test patch and excite the contribution of patches which distribute around the test patch. Similar techniques appeared in image restoration tasks such as image super-resolution \cite{NeighborEmbedding, CVPR2013Yang} and image denoising \cite{CVPR2014Gu}. To further accelerate the synthesis process, we employ principal component analysis (PCA) \cite{Ref5} to reduce the dimension of each patch vector. A graphical outline of the proposed random sampling with locality constraint for face sketch synthesis method (RSLCR) is shown in Fig. \ref{fig:1}. In addition, we proposed a fast version of the proposed method by dropping out some random sampled patches, namely Fast-RSLCR.

\textbf{The contributions of this paper are twofold. \textit{Firstly, an offline random sampling strategy is employed to reduce the time consuming for online neighbor selection.}} In addition, the proposed strategy has stronger scalability than state-of-the-art methods due to the fact that the time-consuming does not depend on the scale of training dataset for our proposed strategy while not the case for other methods. We further imposed locality constraint to the reconstruction weight representation which takes the distinct similarities between the test patch and random sampled patches into consideration. This improves the quality of synthesized sketches. \textbf{\textit{Secondly, both our proposed RSLCR method and its fast version Fast-RSLCR achieve superior performance than state-of-the-art methods in terms of both synthesis performance and synthesis efficiency.}} Specially, our proposed Fast-RSLCR could synthesize a sketch using no more than 1.5 seconds on the Chinese University of Hong Kong (CUHK) face sketch FERET database (CUFSF) under the MATLAB environment, which is the fastest exemplar-based face sketch synthesis method.

In this paper, excepted when noted, a bold lowercase letter represents a column vector, a bold uppercase letter denotes a matrix and regular lowercase and uppercase letters denotes scalars. The rest of this paper is organized as follows. Section \ref{sec:2} introduces the proposed RSLCR method and Fast-RSLCR method. Experimental results and analysis are given in section \ref{sec:3} and section \ref{sec:4} concludes this paper.

\begin{figure}
\centering
\includegraphics[width=0.5\columnwidth]{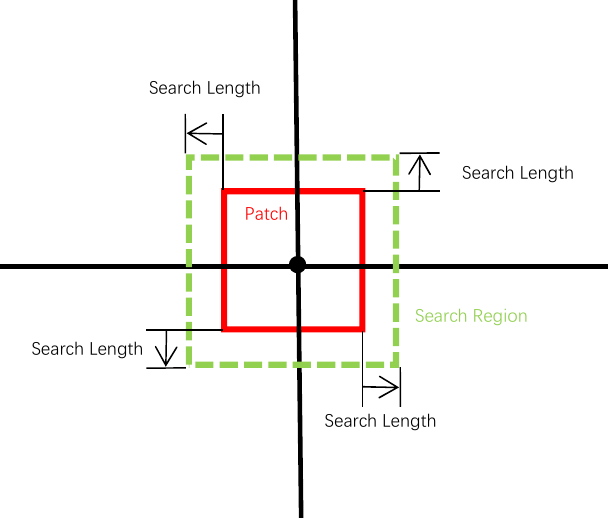}
\caption{Illustration of search region}
\label{fig:2}
\end{figure}

\section{Random Sampling for Face Sketch Synthesis}
\label{sec:2}
In this section, we would introduce how to sample "neighbors" in an offline manner, \textit{i.e.} random sampling training image patches, and then how to represent the test photo patch using these random sampled training photo patches, \textit{i.e.} locality constraint (LCR) based weight representation model.

\subsection{Random Sampling Image Patches}
Supposing there are $M$ pairs of training sketches and training photos which are geometrically aligned according to three points: two eye centers and the mouth center. Each image is cropped to the size of $250\times200$. We first divide these photos and sketches into some patches with even size. There is some overlapping (denoted as $o$) between adjacent patches. As shown in Fig. \ref{fig:1}, each image is divided into $N=40\times31$ patches where there are $r=40$ patches in each column and $c=31$ patches in each row (the patch size $p$ is set to 20 with $o=14$ pixels overlapped between adjacent patches). We reshape each image patch as a column vector. $(i,j)$ denotes the location of the patch at the $i$-th row and the $j$-th column, $i\in\{1,\cdots,r\}, j\in\{1,\cdots,c\}$.

Our target is to generate $N$ clusters of photo-sketch patch pairs corresponding to the $N$ locations $\{(i,j)|i\in\{1,\cdots,r\}, j\in\{1,\cdots,c\}\}$. The most intuitive way is to put patches located at the same position together. However, since images are aligned relying on only three points, there exist misalignments between test photos and training photos, which may result in mismatch during the reconstruction process. To alleviate the influence of misalignment, we enlarge the sampling area to allow more candidate patches bo be sampled as shown in Fig. \ref{fig:2}. Let $s$ denote the search length and then there are $(2s+1)^2$ patches in the search region. Therefore, for each location, we have $(2s+1)^2M$ pairs of patches for sampling. Let $n_{rs}$ denote the number of random sampled patches. In our implementations, we employ the MATLAB function \emph{randperm()} to sample training sketch-photo patch pairs. $\mathbf{X}^{(i,j)}\in\mathbb{R}^{3p^2\times{n_{rs}}}$ and $\mathbf{Y}^{(i,j)}\in\mathbb{R}^{p^2\times{n_{rs}}}$ denote the sampled training photo patches and sketch patches in $(i,j)$-th cluster respectively, $i\in\{1,\cdots,r\}, j\in\{1,\cdots,c\}$.

\begin{table}
\label{algorithm1}
\begin{tabular}{p{0.85\columnwidth}}
\hline
\textbf{Algorithm 1} Random Sampling Image Patches\\
\hline
\textbf{Input}: $p$, $o$, $s$, $n_{rs}$\\
\textit{Step 1}: According to the patch size $p$ and overlap size $o$, compute \\
\qquad \quad $r$, $c$, and the positions of all patches in an image;\\
\textit{Step 2}: Within the search region of each patch position $(i,j)$,\\
\qquad \quad $i\in\{1,\cdots,r\}, j\in\{1,\cdots,c\}$,  random sample $n_{rs}$ pairs \\
\qquad \quad of photo patches $\mathbf{X}^{(i,j)}$ and sketch patches $\mathbf{Y}^{(i,j)}$ in all \\
\qquad \quad training image pairs; \\
\textit{Step 3}: Compute the PCA projection matrix $\mathbf{E}^{(i,j)}$ for each cluster \\
\qquad \quad of training photo patches and project the training photo \\
\qquad \quad patches to the subspace spanned by $\mathbf{E}^{(i,j)}$ as in equation \\
\qquad \quad (\ref{eq:3}).\\
\textbf{Output}: $\mathbf{X}^{(i,j)}$,$\mathbf{Y}^{(i,j)}$, $\mathbf{E}^{(i,j)}$, $i\in\{1,\cdots,r\}, j\in\{1,\cdots,c\}$.\\
\hline
\end{tabular}
\vspace{-3mm}
\end{table}

In order to improve the computation efficiency, we employ PCA to reduce the dimension of training photo patches. 99\% energy is preserved in the projection process. Let $\mathbf{E}^{(i,j)}\in\mathbb{R}^{3p^2\times{D^{(i,j)}}}$ represent the projection matrix and $D^{(i,j)}$ is the reduced dimension. The training photo patches are projected onto the subspace spanned by the column vectors of $\mathbf{E}^{(i,j)}$:
\begin{equation}
\label{eq:3}
\mathbf{X'}^{(i,j)}={\mathbf{E}^{(i,j)}}^T\mathbf{X}^{(i,j)}
\end{equation}
where $\mathbf{X'}^{(i,j)}\in\mathbb{R}^{D^{(i,j)}\times{n_{rs}}}$ is the newly projected training photo patches. For easy of notation, we still use $\mathbf{X}^{(i,j)}$ to denote the projected training photo patches in the following text. Algorithm 1 summarizes the proposed random sampling method.

\subsection{Reconstruction Weight Representation}

Given a test photo $\mathbf{P}\in\mathbb{R}^{250\times{200}}$, it is divided into some patches $\mathbf{x}^{(i,j)}\in\mathbb{R}^{3p^2\times{1}}$ according to the same way for training images, $i\in\{1,\cdots,r\}, j\in\{1,\cdots,c\}$. These patches are projected to the respective subspace obtained in the training phase:
\begin{equation}
\label{eq:4}
\mathbf{x'}^{(i,j)}={\mathbf{E}^{(i,j)}}^T\mathbf{x}^{(i,j)}
\end{equation}
where $\mathbf{x'}^{(i,j)}\in\mathbb{R}^{D^{(i,j)}\times{1}}$ is the projected training photo patch and for easy of notation, we still use $\mathbf{x}^{(i,j)}$ to represent the test photo patch.
In order to take the correlations between different random sampled patches into considerations, we impose a weight to the distances of the test photo patch and random sampled photo patches. Then the reconstruction weight representation model is written as follows:
\begin{equation}
\label{eq:5}
\begin{aligned}
&\min_{\mathbf{w}^{(i,j)}}\|\mathbf{x}^{(i,j)}-\mathbf{X}^{(i,j)}\mathbf{w}^{(i,j)}\|_2^2 + \lambda\|\mathbf{d}^{(i,j)}\odot\mathbf{w}^{(i,j)}\|,\\
&s.t. \mathbf{1}^T\mathbf{w}^{(i,j)} = 1, \forall i\in\{1,\cdots,r\}, \forall j\in\{1,\cdots,c\},
\end{aligned}
\end{equation}
where $\odot$ denotes the element-wise multiplication, $\mathbf{w}^{(i,j)}\in\mathbb{R}^{n_{rs}\times{1}}$ is the weight representation for the test photo patch $\mathbf{x}^{(i,j)}$, $\lambda$ balances the reconstruction error and the locality constraint, and $\mathbf{d}^{(i,j)}\in\mathbb{R}^{n_{rs}\times{1}}$ is the Euclidean distance vector between the test photo patch $\mathbf{x}^{(i,j)}$ and sampled training photo patches $\mathbf{X}^{(i,j)}$. It can be derived that the problem (\ref{eq:5}) has analytical solution:
\begin{equation}
\label{eq:6}
\begin{aligned}
&\mathbf{w'}^{(i,j)}=(\mathbf{C}^{(i,j)}+\lambda{\text{diag}(\mathbf{d}^{(i,j)})})\setminus{\mathbf{1}},\\
&\mathbf{w}^{(i,j)}=\mathbf{w'}^{(i,j)}/\mathbf{1}^T\mathbf{w'}^{(i,j)},
\end{aligned}
\end{equation}
where $\mathbf{1}$ is a column vector of all 1s and its dimension can be determined in the context. $\mathbf{C}^{(i,j)}=(\mathbf{X}^{(i,j)}-\mathbf{1}{\mathbf{x}^{(i,j)}}^T)(\mathbf{X}^{(i,j)}-\mathbf{1}{\mathbf{x}^{(i,j)}}^T)^T$ denotes the data covariance matrix and $\text{diag}(\mathbf{d})$ extends the vector $\mathbf{d}$ into a diagonal matrix. The target sketch patch $\mathbf{y}^{(i,j)}$ is generated from the linear combination of random sampled training sketches weight by the obtained representation vector $\mathbf{w}^{(i,j)}$:
\begin{equation}
\label{eq:7}
\mathbf{y}^{(i,j)}=\mathbf{Y}^{(i,j)}\mathbf{w}^{(i,j)}.
\end{equation}
After obtaining all target sketch patches, they are arranged into a whole sketch with overlapping area averaged.

Since the computation complexity in equation (\ref{eq:6}) mainly depends on the number of random sampled patches, we could further accelerate the proposed RSLCR method by dropping out some random sampled patches in the training phase. Actually we have already computed the distance between the test photo patch and the random sampled training photo patches, we could drop out sampled patches whose distance to the test photo patch are larger. In other words, we could retain $K$ sampled patches whose distance to the test photo patch are among the first $K$ smallest. In comparison to equation (\ref{eq:6}), we only need to update the data matrix $\mathbf{X}^{(i,j)}$ with its subset, \textit{i.e.} $\mathbf{X}_s^{(i,j)}=\mathbf{X}^{(i,j)}(:,\text{idx}(1:K))$ and $\mathbf{Y}_s^{(i,j)}=\mathbf{Y}^{(i,j)}(:,\text{idx}(1:K))$ where idx stores the index to distances between the test photo patch and $n_{rs}$ sampled training photo patches in an ascending order. We call this fast version as Fast-RSLCR. We summarize the proposed RSLCR and Fast-RSLCR algorithm in Algorithm 2.

\begin{table}
\label{algorithm1}
\begin{tabular}{p{0.85\columnwidth}}
\hline
\textbf{Algorithm 2} RSLCR \& Fast-RSLCR\\
\hline
\textbf{Input}: $\mathbf{P}$, $p$, $o$, $K$\\
\textit{Step 1}: According to the patch size $p$ and overlap size $o$, divide \\
\qquad \quad $\mathbf{P}$ into patches $\mathbf{x}^{(i,j)}$,$i\in\{1,\cdots,r\}$,$j\in\{1,\cdots,c\}$;\\
\textit{Step 2}: For $i\in\{1,\cdots,r\}$ \\
\qquad \qquad \ For $j\in\{1,\cdots,c\}$ \\
\textit{Step 3} \qquad RSLCR: Compute $\mathbf{C}^{(i,j)}$ with $\mathbf{X}^{(i,j)}$;\\
\qquad \qquad \quad Fast-RSLCR: Compute $\mathbf{C}^{(i,j)}$ with $\mathbf{X}_s^{(i,j)}$;\\
\textit{Step 4}: \qquad  Compute $\mathbf{w}^{(i,j)}$ as in equation (\ref{eq:6});\\
\textit{Step 5}: \qquad RSLCR: $\mathbf{y}^{(i,j)}=\mathbf{Y}^{(i,j)}\mathbf{w}^{(i,j)}$;\\
\qquad \qquad \quad Fast-RSLCR: $\mathbf{y}^{(i,j)}=\mathbf{Y}_s^{(i,j)}\mathbf{w}^{(i,j)}$;\\
\textit{Step 6}: Arrange all target sketch patches into a whole sketch with\\
\qquad \quad overlapping area averaged.\\
\textbf{Output}: the target sketch $\mathbf{S}$.\\
\hline
\end{tabular}
\vspace{-3mm}
\end{table}

\section{Experimental Results and Analysis}
\label{sec:3}
\begin{figure}
\centering
\includegraphics[width=0.5\columnwidth]{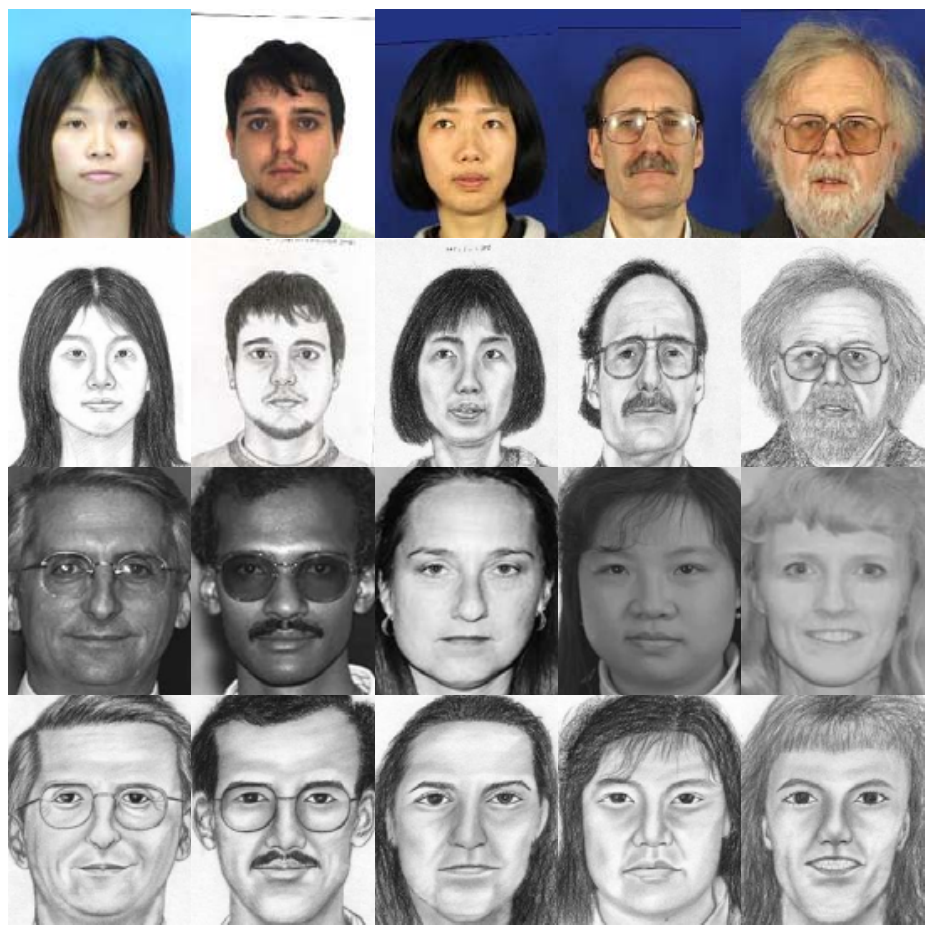}
\caption{ {Example face sketch-photo pairs in the CUFS database (the first two rows) and the CUFSF database (the last two rows). The first and the third row are face photos and the second and the last rows are corresponding face sketches drawn by the artist. The first person, second person and the last three persons at the first two rows are from the CUHK student database, the AR database and the XM2VTS database respectively.}}
\label{fig:3}
\end{figure}

Experimental results are conducted to illustrate the efficiency and effectiveness of the proposed RSLCR method and Fast-RSLCR method. Two public available database are used: the CUHK face sketch database (CUFS) \cite{Ref13} and the CUFSF database \cite{Ref31}. The CUFS database consists of face photos from three databases: the CUHK student database \cite{Ref2} (188 persons), the AR database \cite{Ref23} (123 persons) and the XM2VTS database \cite{Ref24} (295 persons). Persons in the XM2VTS database are different in ages, skins (races) and hair styles. The CUFSF database includes 1194 persons from the FERET database \cite{Ref32}. There are one face photos and one face sketch drawn by the artist for each person in both CUFS and CUFSF databases. Face photos in the CUFSF database are with lighting variation and sketches are with shape exaggeration. All these face photos and sketches are geometrically aligned relying on three points: two eye centers and the mouth center and they are cropped to the size of $250\times{200}$. Fig. \ref{fig:3} gives some examples from these two databases.

In the following context, we would first discuss the experimental settings (parameter settings) for our proposed method on the CUHK student database. Afterwards, under the experimental settings, we perform face sketch synthesis on the CUFS database and the CUFSF database to subjectively illustrate the superiority of the proposed RSLCR method and the Fast-RSLCR method compared with state-of-the-arts. Then time consumption has discussed. Subsequently, objective statistic experiments (objective image quality assessment and face recognition) are conducted to indirectly validate the superiority of proposed methods.

\subsection{ {Discussion on Experimental Settings}}
We employ the CUHK student database \cite{Ref2} to perform parameter adjusting in this sub-section. 88 pairs of face photo-sketch are taken as the training set and the rest 100 pairs of face photo-sketch are taken for validation. To objectively assess the quality of synthesized sketches under different experimental settings, structural similarity index metric (SSIM) \cite{Ref25} is adopted as the evaluation criterion. The 100 sketches drawn by the artist in the validation set are taken as the reference image and 100 photos in the validation set are taken as the test image for face sketch synthesis. Under each experimental setting, the average SSIM score of 100 synthesized sketches are taken as the final evaluation value.

\begin{figure}
\centering
\subfigure[]{
\label{fig:subfig4a}
\includegraphics[width=0.3\columnwidth]{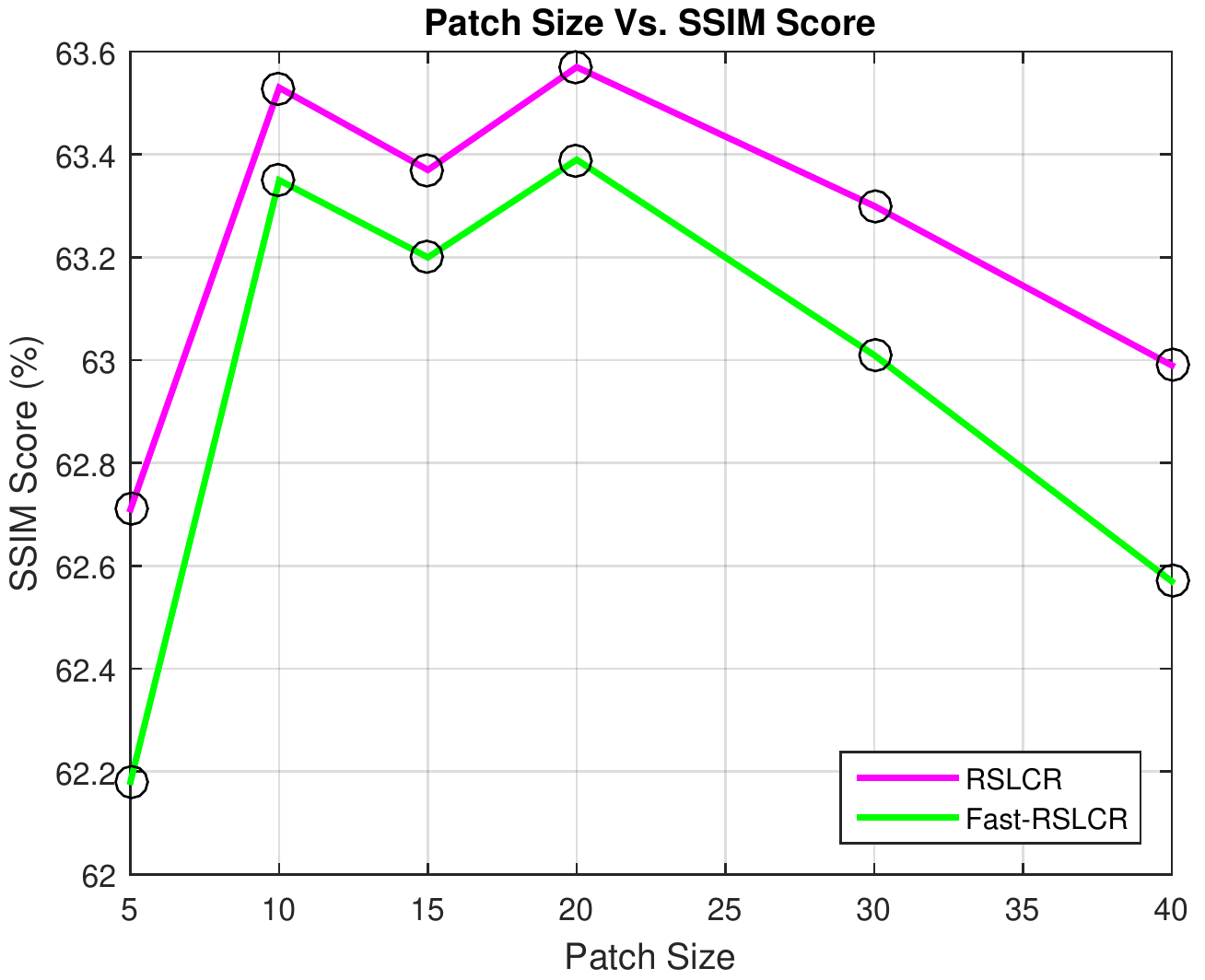}
}
\subfigure[]{
\label{fig:subfig4b}
\includegraphics[width=0.3\columnwidth]{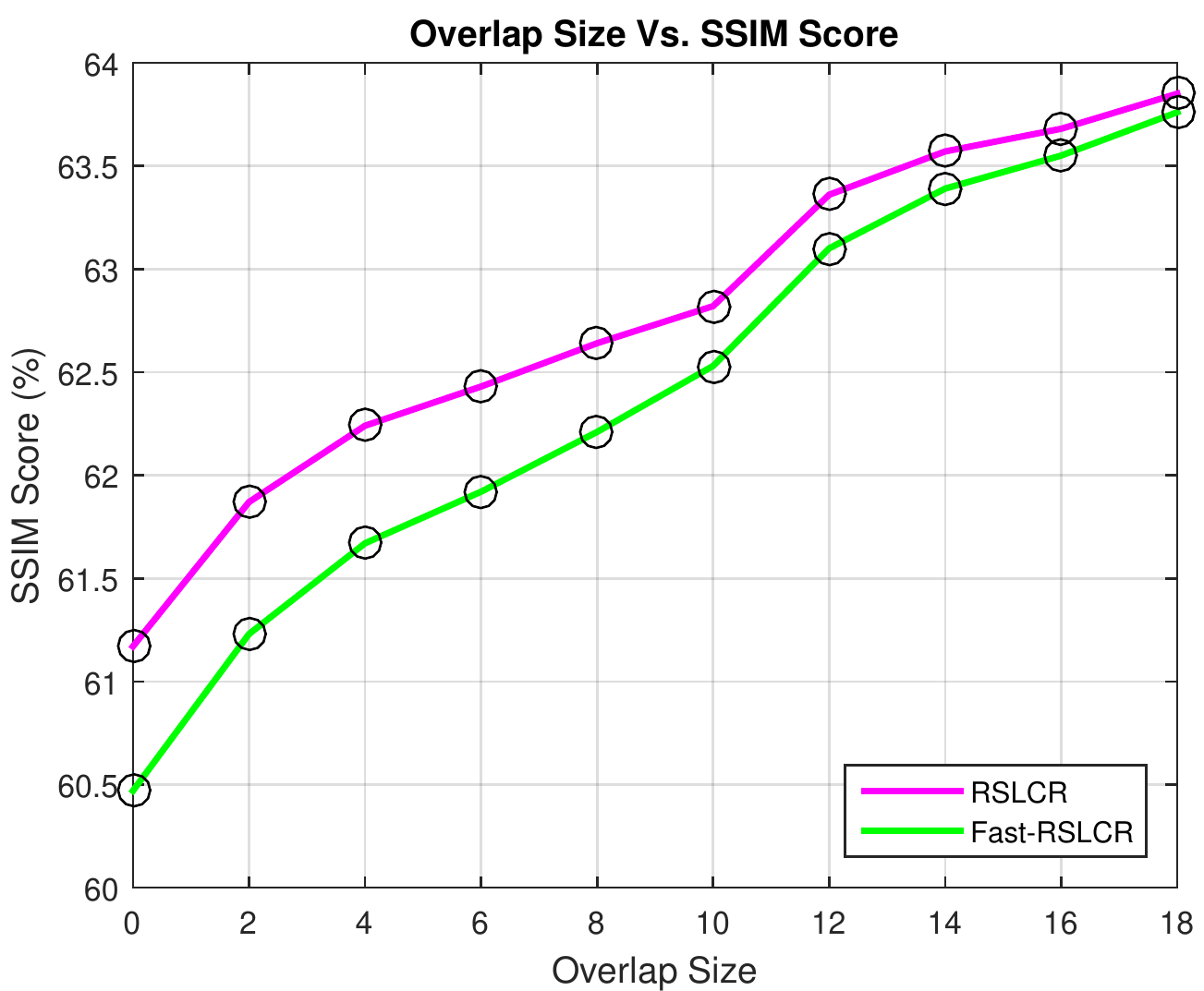}
}
\subfigure[]{
\label{fig:subfigc}
\includegraphics[width=0.3\columnwidth]{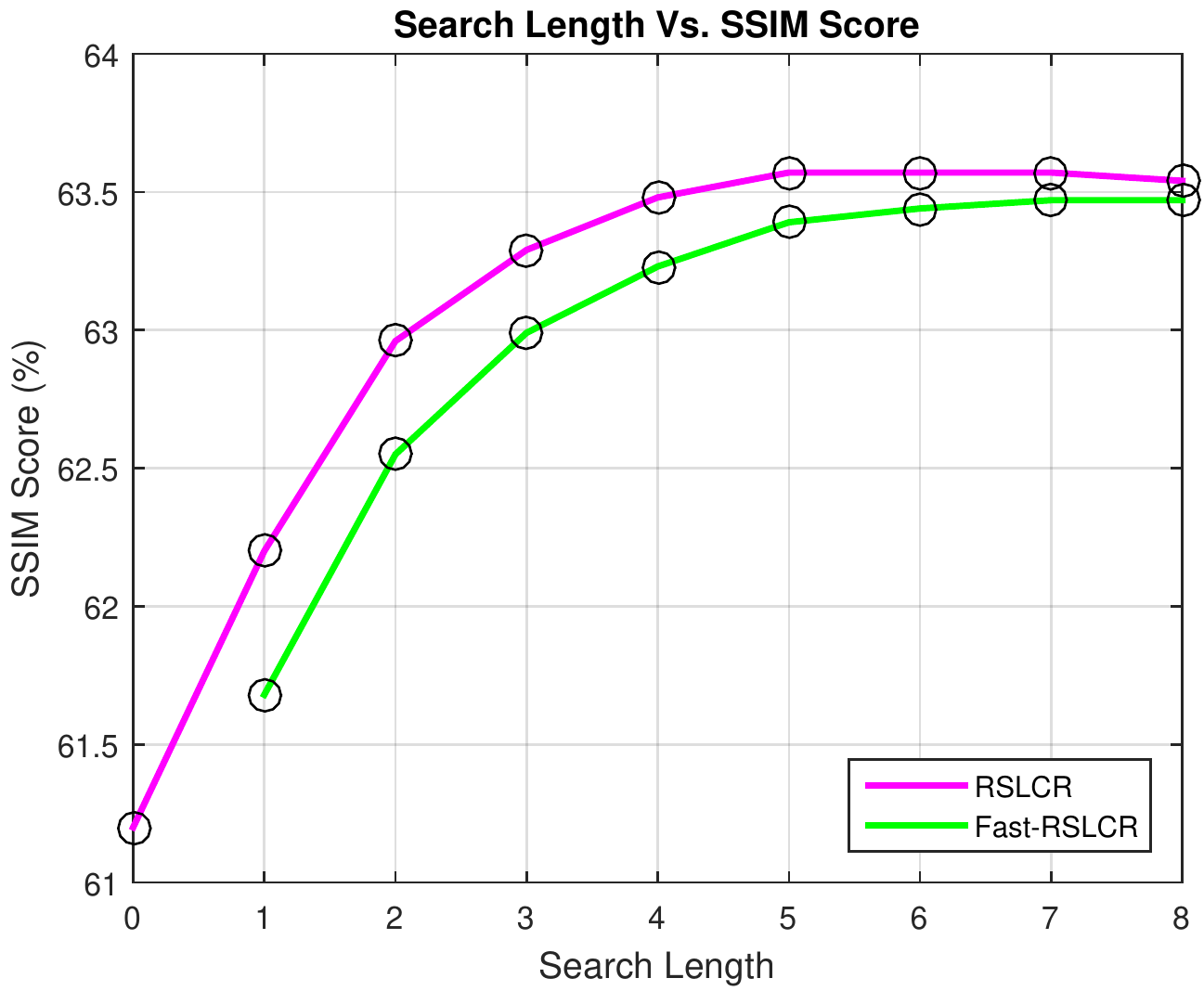}
}
\subfigure[]{
\label{fig:subfigd}
\includegraphics[width=0.3\columnwidth]{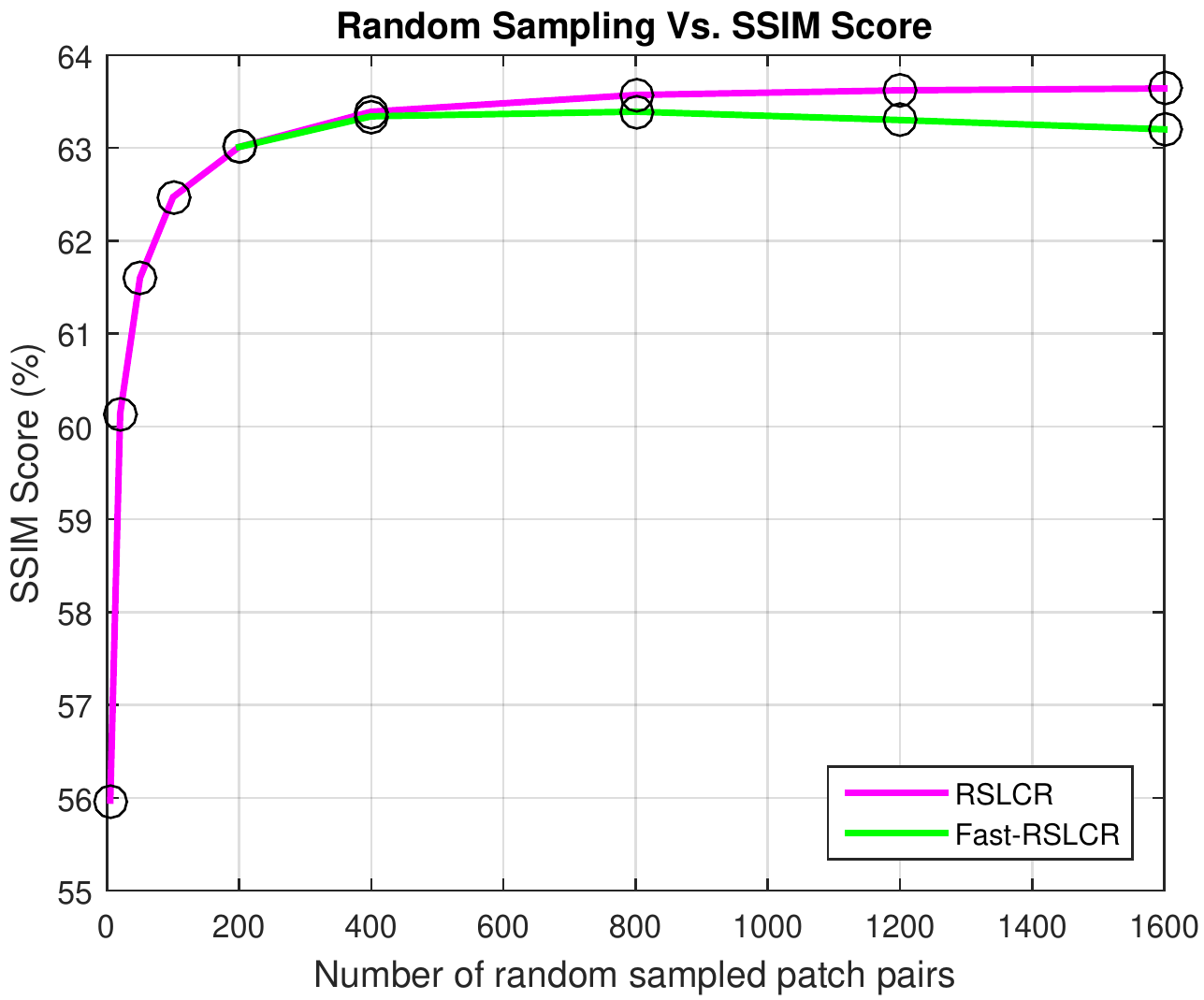}
}
\subfigure[]{
\label{fig:subfige}
\includegraphics[width=0.3\columnwidth]{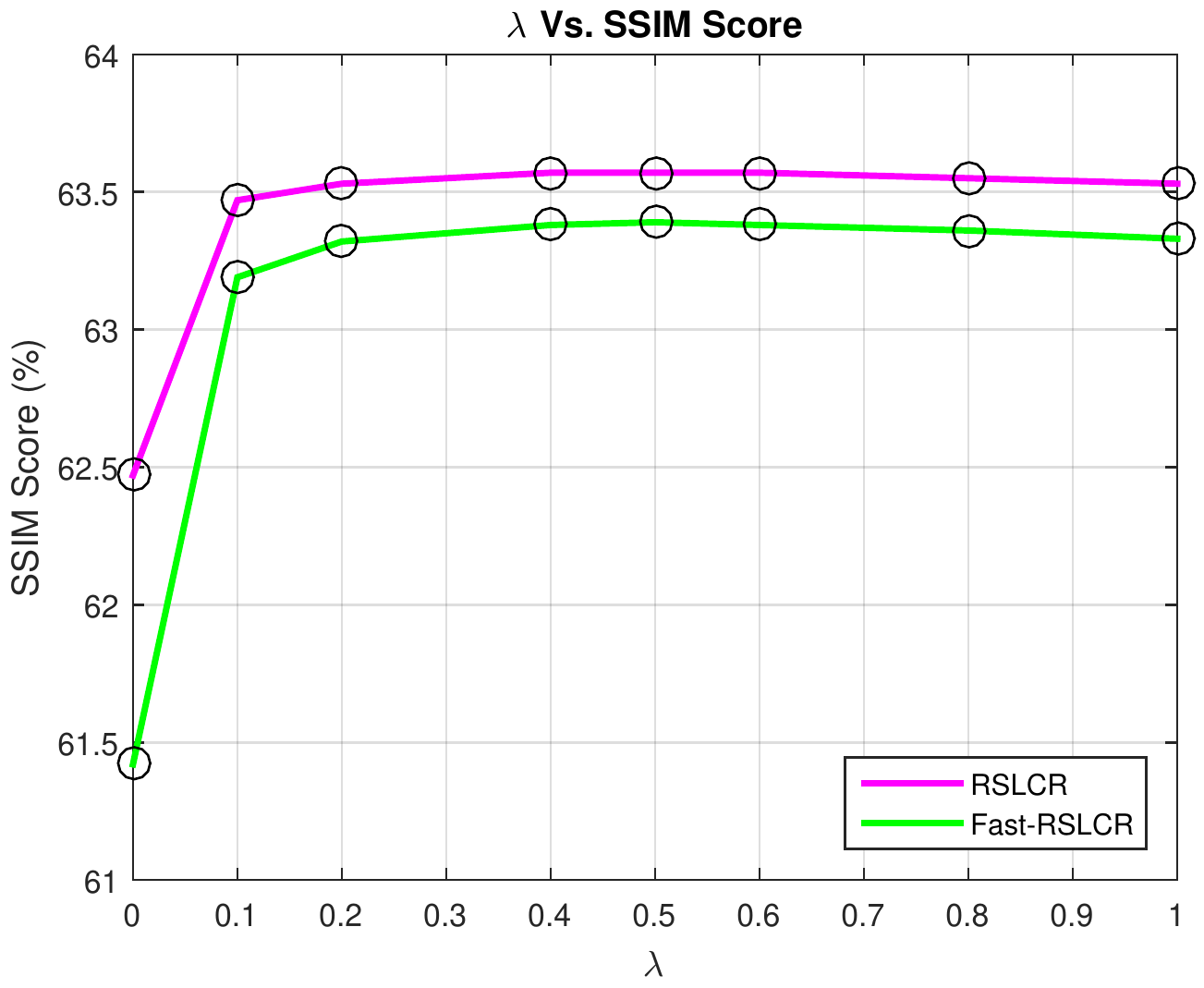}
}
\subfigure[]{
\label{fig:subfigf}
\includegraphics[width=0.3\columnwidth]{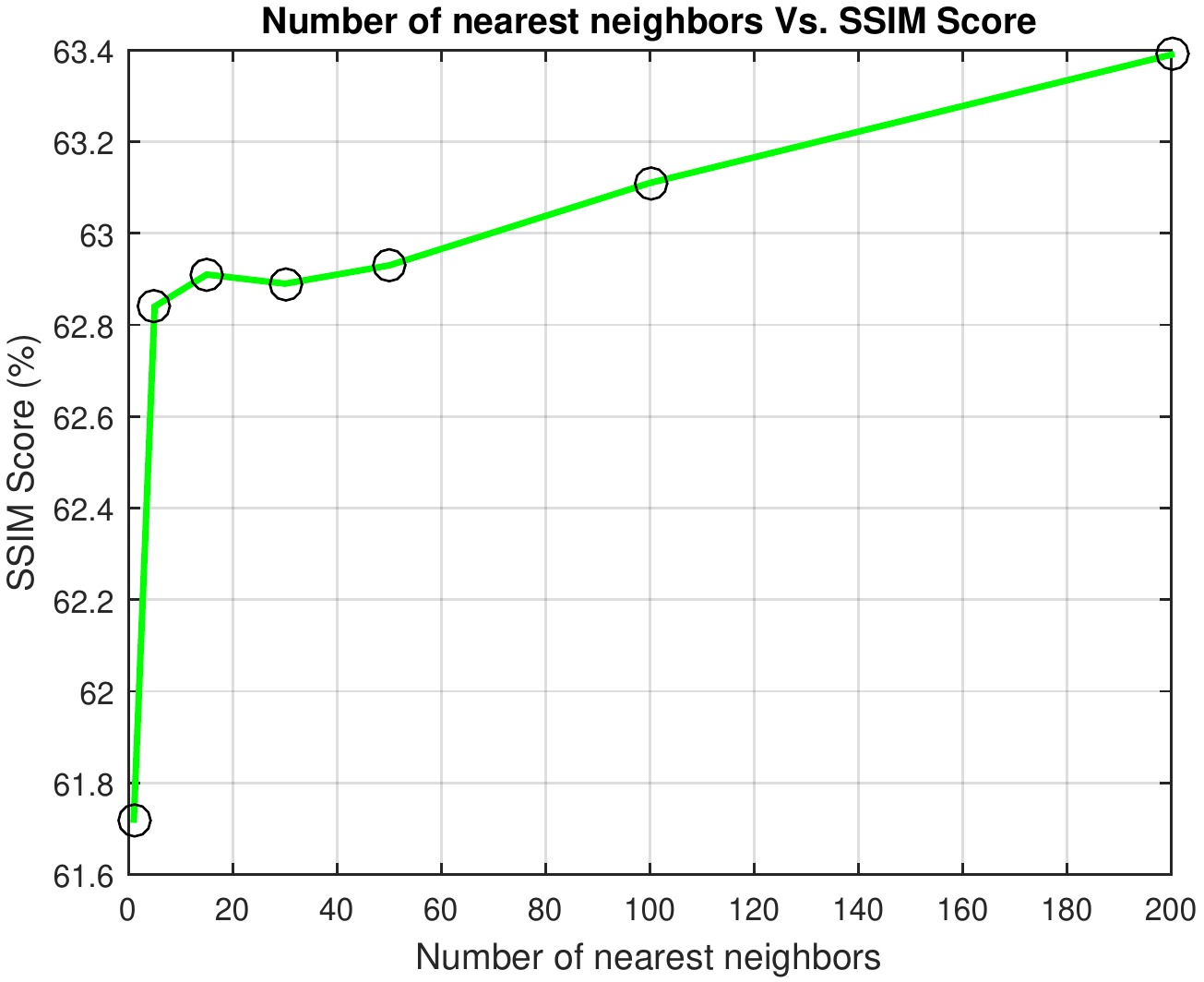}
}
\caption{Statistics of SSIM scores under different parameter settings: (a) patch size, (b) overlap size, (c) search length, (d) number of random sampled patch pairs, (e) $lambda$, (f) number of nearest neighbors for Fast-RSLCR.}
\label{fig:4}
\end{figure}

There are five parameters (\textit{i.e.} patch size $p$, overlap size $o$, search length $s$, the number of random sampled patches $r_{ns}$ and the trade-off parameter $\lambda$) for the proposed RSLCR method and one additional parameter (\textit{i.e.} the number of selected neighbors) for Fast-RSLCR. All experiments are conducted using MATLAB R2015a on Windows 7 system with i7-4790 3.6G CPU. Fig. \ref{fig:4} presents the SSIM scores against different parameter settings.

\subsubsection{ {Patch Size}} We set patch size to 5, 10, 15, 20, 30 and 40 respectively and keep the overlap size as 70\% of the patch size. It can be seen from Fig. \ref{fig:subfig4a} that both RSLCR and Fast-RSLCR achieves the highest SSIM score when the patch size is 20. When the patch size is 10, it has very close performance with the patch size 20. However, smaller patch size means more patches to be synthesized and hence it consumes much more time (78.84 seconds vs. 18.79 seconds for RSLCR and 21.07 seconds vs. 1.82 seconds for Fast-RSLCR).

\subsubsection{ {Overlap Size}} Given the patch size being 20, we set the overlap size to different values: 0, 2, 4, 6, 8, 10, 12, 14, 16 and 18. From Fig. \ref{fig:subfig4b} it can be seen that with the increase of the overlap size, SSIM scores for both RSLCR and Fast-RSLCR are also increasing. However, the time-consuming is also growing rapidly. In our following experiments, we set the overlap size as the trade-off value 14.

\subsubsection{ {Search Length}} The search length is used in the training stage and hence it does not affect the test phase. From Fig. \ref{fig:subfigc} it can be seen that though the SSIM score of Fast-RSLCR always grows with the increase of the search length, the SSIM score of RSLCR grows at first and then goes down. This is because with the increase of the search length, some outliers may be sampled and these outliers would bring noise to the RSLCR method. However, Fast-RSLCR does not subject to these outliers because it selects the $K$ nearest neighbors among random sampled patches which could filter these outliers. The search length is set to 5 in our experiments.

\subsubsection{ {Random Sampling}} Fig. \ref{fig:subfigd} presents the SSIM score corresponding to different number of random sampled training face photo-sketch patches. It can be seen that generally the SSIM score grows with the increase of random sampling number for the RSLCR method. However, it begins to decrease when the number is bigger than 800 for the Fast-RSLCR method. We set it to 800 in our experiments.

\subsubsection{ {Regularization Parameter}}$\lambda$ It should be noted that when $\lambda=0$ (refer to equation (\ref{eq:5})), the locality constraint has no contribution to the reconstruction weight computation. Then equation (\ref{eq:5}) reduces to the LLE model in equation (\ref{eq:1}). The difference is that the entries of the data matrix $\mathbf{X}$ for the LLE method \cite{Ref6} is selected through $K$-NN while they are random sampled for RSLCR and Fast-RSLCR. From Fig. \ref{fig:subfige}, on the one hand, it can be seen that when $\lambda=0$, SSIM scores for RSLCR and Fast-RSLCR are 0.6250 and 0.6150 respectively compared with 0.5990 of the LLE method \cite{Ref6}. In addition, RSLCR and Fast-RSLCR run much faster than the LLE method (18.79 seconds for RSLCR, 1.82 seconds for Fast-RSLCR, and 536.34 seconds for LLE). This illustrates the effectiveness of the proposed random sampling strategy. On the other hand, when $\lambda=0.5$, RSLCR and Fast-RSLCR achieve the best performance among all $\lambda$ values, which is much larger than SSIM socres for $\lambda=0$. This validates that the locality constraint does help to improve the performance. $\lambda$ is set to 0.5 in this paper.

\subsubsection{ {Number of Nearest Neighbors for Fast-RSLCR}} $K$-NN is conducted in Fast-RSLCR to improve the computation efficiency. Fig. \ref{fig:subfigf} presents the SSIM score against different number of nearest neighbors. Generally it grows with the increase of the number. To comprehensively comprise the time consuming and the SSIM socre, it is set to 200 in our experiments. From Fig. \ref{fig:subfigd} it can be seen that directly random sampling 200 patches achieves an SSIM score of 0.6301 (at the time cost of 1.82s) while our proposed Fast-RSLCR (also use 200 sampled patches) could achieve an SSIM score of 0.6339 (at the time cost of 1.89s). It demonstrates the effectiveness of the proposed Fast-RSLCR method.

\subsubsection{ {Random Sampling Searching Vs. Accelerated Nearest Neighbor Searching}}  
\begin{figure}
\centering
\includegraphics[width=1\columnwidth]{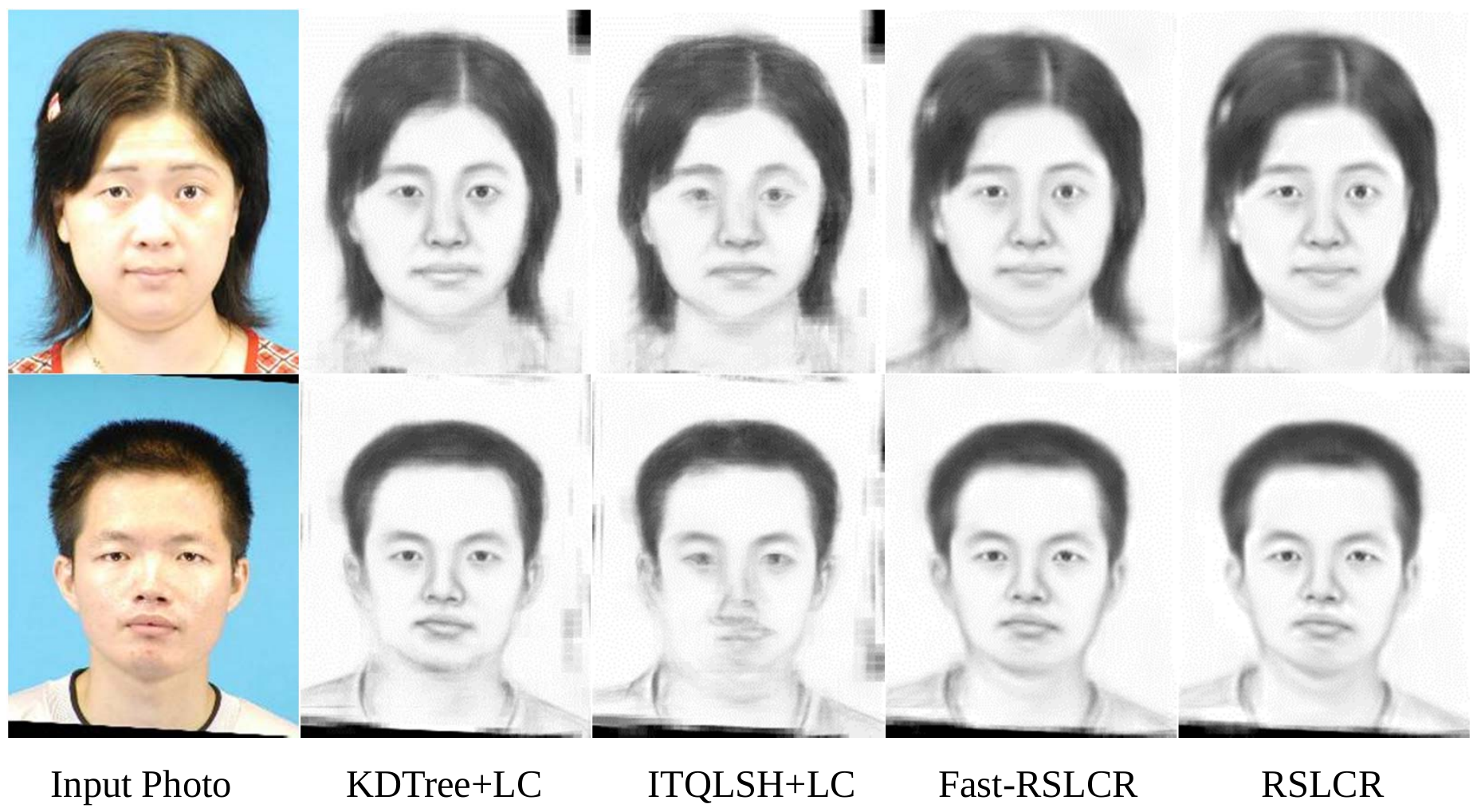}
\caption{ {Synthesized sketches by KDTree-LC, ITQLSH-LC, Fast-RSLCR, and RSLCR method respectively.}}
\label{fig:45}
\end{figure}

{In order to illustrate the effectiveness of the proposed the offline random sampling searching strategy in comparison to accelerated nearest neighbor (ANN) searching strategies, we utilize two kinds of ANN approaches for online neighbor searching: iterative quantization based locality-sensitive hashing (ITQLSH) \footnote{The MATLAB/C++ mixed source codes are download from the website: \url{https://github.com/RSIA-LIESMARS-WHU/LSHBOX}} \cite{ITQLSH} and the KDTree method \footnote{We use the open source implementation (MATLAB/C++ mixed programming) of this method in VLFeat: \url{http://www.vlfeat.org/}} \cite{KDTree}. We substitute the random sampling strategy in our proposed method with these two KNN searching method and they are denoted as ITQLSH-LC and KDTree-LC respectively. Table \ref{tab:0} gives the effect of the number of nearest neighbors on these two methods. It can be seen that when the number of neighbors is 20, these two methods achieve the best performance. Table \ref{tab:00} shows the comparison between the proposed methods with these two ANN based methods. Fig. \ref{fig:45} presents the synthesized sketches by four different methods. Table \ref{tab:00} and Fig. \ref{fig:45} illustrate that the random sampling strategy outperforms ITQLSH and KDTree in terms of both time consumption and image quality.}

\begin{table*}
\centering
\caption{ {The Effect of the Number of Nearest Neighbors on ITQLSH-LC and KDTree in terms of SSIM Score (\%) and Time Consumption (the Value in the Bracket Is in Seconds)}}
\label{tab:0}
 \scalebox{0.85}{
\begin{tabular}{cccccccc}
\hline
Number of Neighbors  & 5   & 20   & 50  & 100 & 200  & 400 & 800 \\
\hline
 KDTree-LC &  60.72 (434.89) & \textbf{61.27} (436.18)& 61.13 (437.03) & 60.72 (439.77)& 60.36 (462.63)& 60.34 (474.85)& 60.50 (501.57)\\
 ITQLSH-LC &  58.80 (412.16) & \textbf{59.30} (421.09)& 58.96 (432.16) & 58.58 (437.24)& 58.53 (439.78)& 58.65 (443.14)& 58.76 (461.01)\\
\hline
\end{tabular}}
\end{table*}

\begin{table}
\centering
\caption{ {Comparisons between Two ANN based Methods and the Proposed Methods in terms of SSIM Score (\%) and Time Consumption (Seconds)}}
\label{tab:00}
 {
\begin{tabular}{ccccc}
\hline
Method  & KDTree-LC  & ITQLSH-LC   & RSLCR & Fast-RSLCR \\
\hline
SSIM & 61.27  &  59.30 & 63.57 & 63.39 \\
Time & 436.18 & 421.09 & 18.79 & 1.82  \\
\hline
\end{tabular}}
\end{table}

\subsubsection{ {Locality Constraint (LC)}}  {Locality constraint is firstly proposed for image classification and it shows comparable performance with sparse constraint \cite{Ref30}. The non-local similarity constraint as a regularization term in \cite{Ref10} is also very similar with the locality constraint.  Table \ref{tab:000} compares the performance of different face sketch synthesis methods with or without locality constraint. It can be found that locality constraint improves the performance of our proposed Fast-RSLCR method a lot and it also improves the LLE method and the RSLCR method. However, this is not the case for the MWF method. This is because the locality constraint is implicitly embedded in the neighboring constraint of the MWF model (see the second term of equation (\ref{eq:2})). In this paper, locality constraint is utilized to distinguish the random sampled patches since these patches may be distributed in a scattered way, \textit{i.e.} it is especially appropriate for random sampling based methods where the random sampled patches may distribute far away from each other.}
\begin{table}
\centering
\caption{ {The Effect of Locality Constraint (in terms of SSIM Score (\%)) on Different Face Sketch Synthesis Methods}}
\label{tab:000}
 {
\begin{tabular}{ccccc}
\hline
Method  & LLE  & MWF & Fast-RSLCR & RSLCR\\
\hline
With LC & 61.00  &  62.31 & 63.39 & 63.57 \\
Without LC & 59.97 & 62.31 & 61.42 & 62.47  \\
\hline
\end{tabular}}
\end{table}

\subsection{Face Sketch Synthesis}
\label{subsec:fss}

\begin{figure*}
\centering
\includegraphics[width=1\columnwidth]{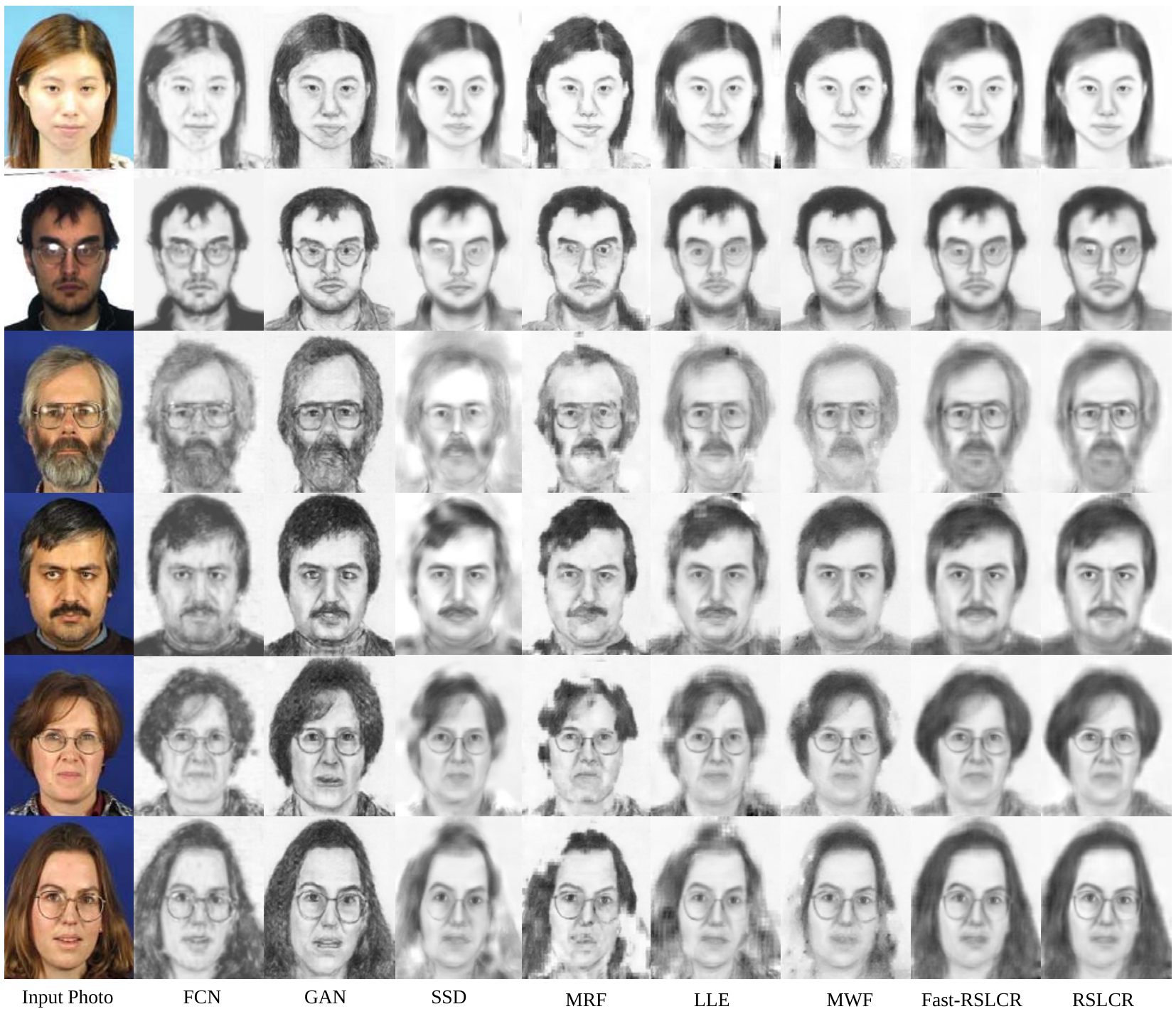}
\caption{ {Synthesized sketches on the CUFS database by FCN \cite{FCN}, GAN \cite{cGAN}, SSD \cite{Ref1}, MRF \cite{Ref13}, LLE \cite{Ref6}, MWF \cite{Ref15}, the proposed Fast-RSLCR and RSLCR respectively. Face photos in the first two rows are from the CUHK student database and the AR database respectively. The last four photos are from the XM2VTS database.}}
\label{fig:5}
\end{figure*}

After the experimental setting for parameters, we set the patch size $p=20$, overlap size $o=14$, search length $s=5$, the number for random sampling $n_{rs}=800$, the regularization parameter $\lambda=0.5$, the number of nearest neighbors for Fast-RSLCR $K=200$. For the CUHK student database, 88 pairs of face photo-sketch are taken for training and the rest for testing (the data has been partitioned in this database). For the AR database, we randomly choose 80 pairs for training and the rest 43 pairs for testing. For the XM2VTS database, we randomly choose 100 pairs for training and the rest 195 pairs for testing.  {Six state-of-the-art methods are compared: the FCN method \cite{FCN}, the GAN method \cite{cGAN}\footnote{ {The source codes are availabe online: \url{https://github.com/phillipi/pix2pix}}}, the LLE method \cite{Ref6}, the SSD method \cite{Ref1}, the MRF method \cite{Ref13}, and the MWF method \cite{Ref15}.} All synthesized sketches by the SSD method and the MWF method are generated from the source codes provided by the authors. For the MRF method, we use the codes from the implementation provided by authors of SSD \cite{Ref1}\footnote{The source codes for both the MRF method and the SSD method are available online: \url{http://www.cs.cityu.edu.hk/~yibisong/eccv14/index.html}}.  {Results of the LLE method and the FCN method are based on our implementations}\footnote{Available online: \url{http://www.ihitworld.com/RSLCR.html}. On this project website, we also release the source codes of both our proposed methods and the evaluation codes (objective image quality assessment codes and face recognition codes).}. The full list of synthesized sketches (both of our methods and all four compared methods) on these two databases is available on our project website.

\begin{figure*}
\centering
\includegraphics[width=1\columnwidth]{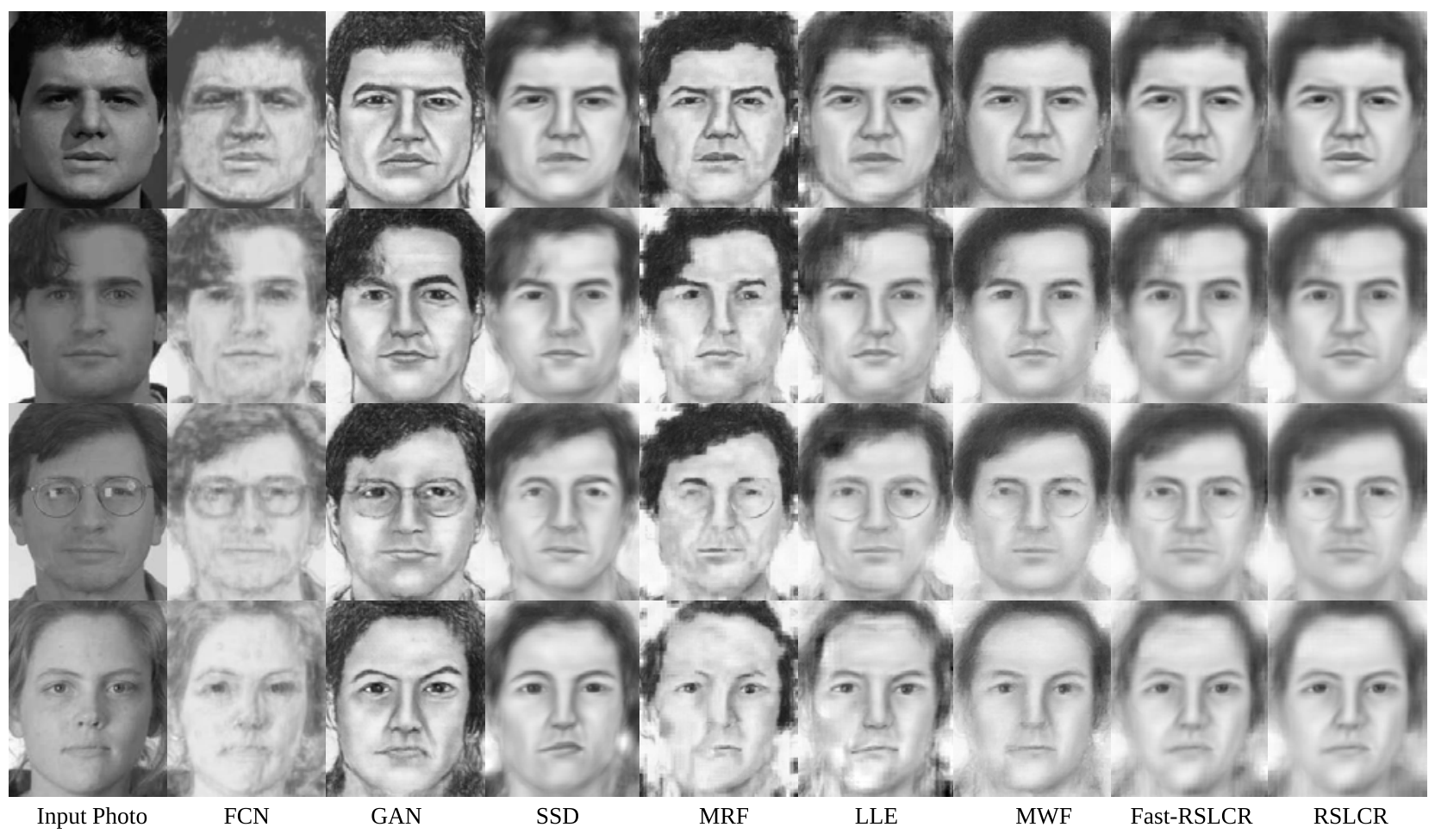}
\caption{ {Synthesized sketches on the CUFSF database by FCN \cite{FCN}, GAN \cite{cGAN}, SSD \cite{Ref1}, MRF \cite{Ref13}, LLE \cite{Ref6}, MWF \cite{Ref15}, the proposed Fast-RSLCR and RSLCR respectively.}}
\label{fig:6}
\end{figure*}

Fig. \ref{fig:5} presents some synthesized face sketches from different methods on the CUFS database. It can be seen that the proposed RSLCR method and the Fast-RSLCR method could generate fine textures (\textit{e.g.} hair region) and structures (\textit{e.g.} glasses). This is because more candidate patches in our proposed methods are effectively incorporated through random sampling and locality constraint. Synthesized sketches on photos from the XM2VTS database generated by SSD, MRF, LLE, and MWF are less satisfying than photos from the CUHK student database and the AR database due to the fact that there are more variations such as aging, race, and hair styles on faces of the XM2VTS database. However, RSLCR and Fast-RSLCR achieve much better performance than these four comared methods and afford comparable performance on face photos from three different databases. This illustrates the robustness of the proposed methods.

We have also investigated the robustness of the proposed methods against shape exaggeration and illumination variations on the CUFSF database. We randomly choose 250 face photo-sketch pairs for training and the rest 944 pairs for test. Fig. \ref{fig:6} shows the synthesized results on this database by various methods. It is shown that there are some deformations on synthesized sketches by SSD and MRF, specially for the mouth area. In addition, our proposed Fast-RSLCR and RSLCR method could handle glasses with reflect light well while other methods cannot (see the third row of Fig. \ref{fig:6}).

\subsection{Time Consumption}

Given the experimental settings in \ref{subsec:fss}, we count the time consumption for the proposed methods. Table \ref{tab:1} compares the time consuming for different methods on different databases. There are 88, 80, 100, and 250 training photo-sketch pairs for the CUHK student, AR, XM2VTS and FERET database respectively. It can be seen from the table that time consumptions for SSD, MRF, LLE, and MWF are proportional to the scale of the training set because these methods search neighbors by traversing the whole training dataset. However, our proposed RSLCR and Fast-RSLCR method costs comparable time on these four databases. This validates the stronger scalability of the proposed RSLCR framework. Moreover, it can be seen that RSLCR has comparable or even less time consuming compared with state-of-the-art methods. Our proposed Fast-RSLCR is the most efficient method among all methods. It requires less than 1.5 seconds to synthesize a sketch on the CUHK FERET database, which is dozens of times faster than state-of-the-art methods.

\begin{figure*}
\centering
\subfigure[]{
\label{fig:subfiga}
\includegraphics[width=0.45\columnwidth]{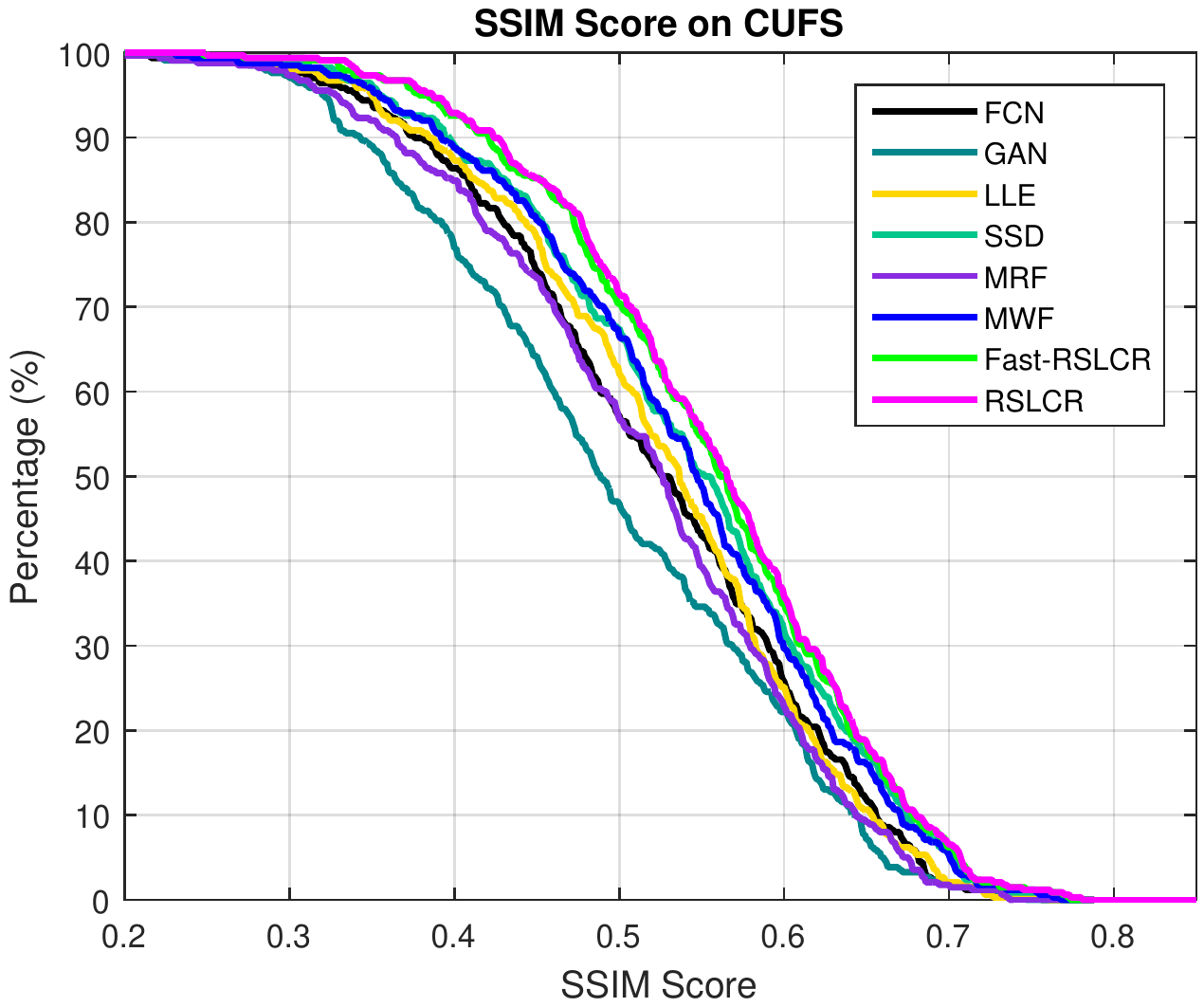}
}
\subfigure[]{
\label{fig:subfigb}
\includegraphics[width=0.45\columnwidth]{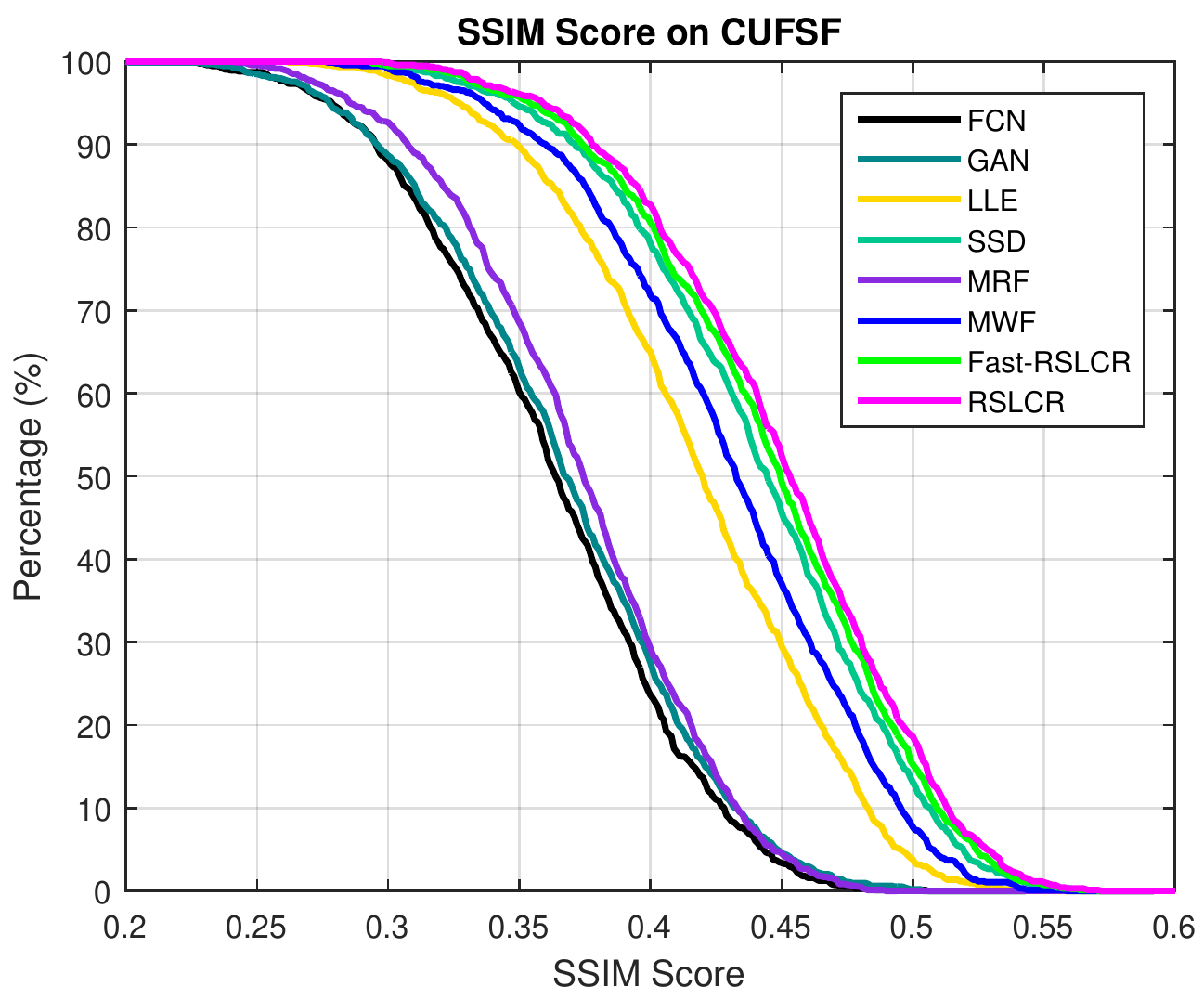}
}
\caption{ {Statistics of SSIM scores on (a) the CUFS database and (b) the CUFSF database.}}
\label{fig:7}
\end{figure*}

\begin{table*}
\centering
\caption{Average time consumption (seconds) to generate one sketch on different databases}
\label{tab:1}
\begin{tabular}{ccccccc}
\hline
Methods              & SSD   & MRF   & LLE    & MWF   & RSLCR  & Fast-RSLCR \\
\hline
Programming language & C++   & C++   & MATLAB & C++   & MATLAB & MATLAB \\
CUHK Student         & 4.50  & 8.60  & 536.34 & 16.10 & 18.79  & \textbf{1.82}   \\
AR                   & 4.10  & 8.40  & 496.47 & 15.33 & 19.10  & \textbf{1.73}   \\
XM2VTS               & 5.10  & 10.4  & 642.50 & 18.80 & 18.14  & \textbf{2.36}   \\
CUHK FERET           & 11.60 & 24.25 & 1591.95& 45.20 & 17.66  & \textbf{1.44}   \\
\hline
\end{tabular}
\end{table*}

\begin{table*}
\centering
\caption{ {Average SSIM score (\%) on the CUFS database and the CUFSF database}}
\label{tab:2}
 {
\begin{tabular}{cccccccccc}
\hline
& FCN \cite{FCN} & GAN \cite{cGAN} & SSD\cite{Ref1} & MRF\cite{Ref13} & LLE\cite{Ref6} & MWF\cite{Ref15} & Fast-RSLCR & RSLCR \\
\hline
CUFS (\%) & 52.14& 49.39 & 54.20 & 51.32 &  52.58 & 53.93 & \textbf{55.42} & \textbf{55.72} \\
CUFSF (\%) & 36.22& 36.65 & 44.09  & 37.34 & 41.76 & 42.99 & \textbf{44.56} & \textbf{44.96} \\
\hline
\end{tabular}}
\end{table*}

\subsection{Objective Image Quality Assessment}

We utilize SSIM to evaluate the quality of synthesized sketches by different methods on CUFS and CUFSF. There are 338 ($100+43+195$) and 944 synthesized sketches for each method generated from the CUFS database and CUFSF database respectively. Fig. \ref{fig:7} gives the statistics of SSIM scores on these two databases respectively. The horizontal axis labels the SSIM score from 0 to 1. The vertical axis means the percentage of synthesized sketch whose SSIM scores are not smaller than the score marked on the horizontal axis.  Table \ref{tab:2} presents the average SSIM score on the CUFS and CUFSF database respectively.

It can be seen from Fig. \ref{fig:7} and table \ref{tab:2} that both Fast-RSLCR and RSLCR outperform four other state-of-the-art methods. Comparable performance is achieved for SSD and MWF on the CUFS database but SSD outperforms MWF on the CUFSF database which illustrates SSD could handle faces with illumination variations better than the MWF method.

\begin{table*}
\centering
\caption{ {NLDA face recognition accuracy (\%) based on synthesized sketches from the CUFS database and the CUFSF database}}
\label{tab:3}
 {
\begin{tabular}{ccccccccc}
\hline
Methods  & FCN& GAN & LLE & SSD & MRF & MWF & Fast-RSLCR & RSLCR  \\
\hline
CUFS (\%) & 96.49 (131) & 93.48 (139) & 91.12 (148) & 90.24 (149) & 87.29 (149) & 92.13 (149) & \textbf{98.35} (121) & \textbf{98.38} (133)\\
CUFSF (\%) & 69.80 (237) & 71.44 (164) & 61.76 (274) & 70.92 (266) & 46.03 (223) & 74.15 (299) & \textbf{73.41} (287) & \textbf{75.94} (296) \\
\hline
\end{tabular}}
\end{table*}


\subsection{Face Sketch Recognition}

\begin{figure*}
\centering
\subfigure[]{
\label{fig:subfiga}
\includegraphics[width=0.45\columnwidth]{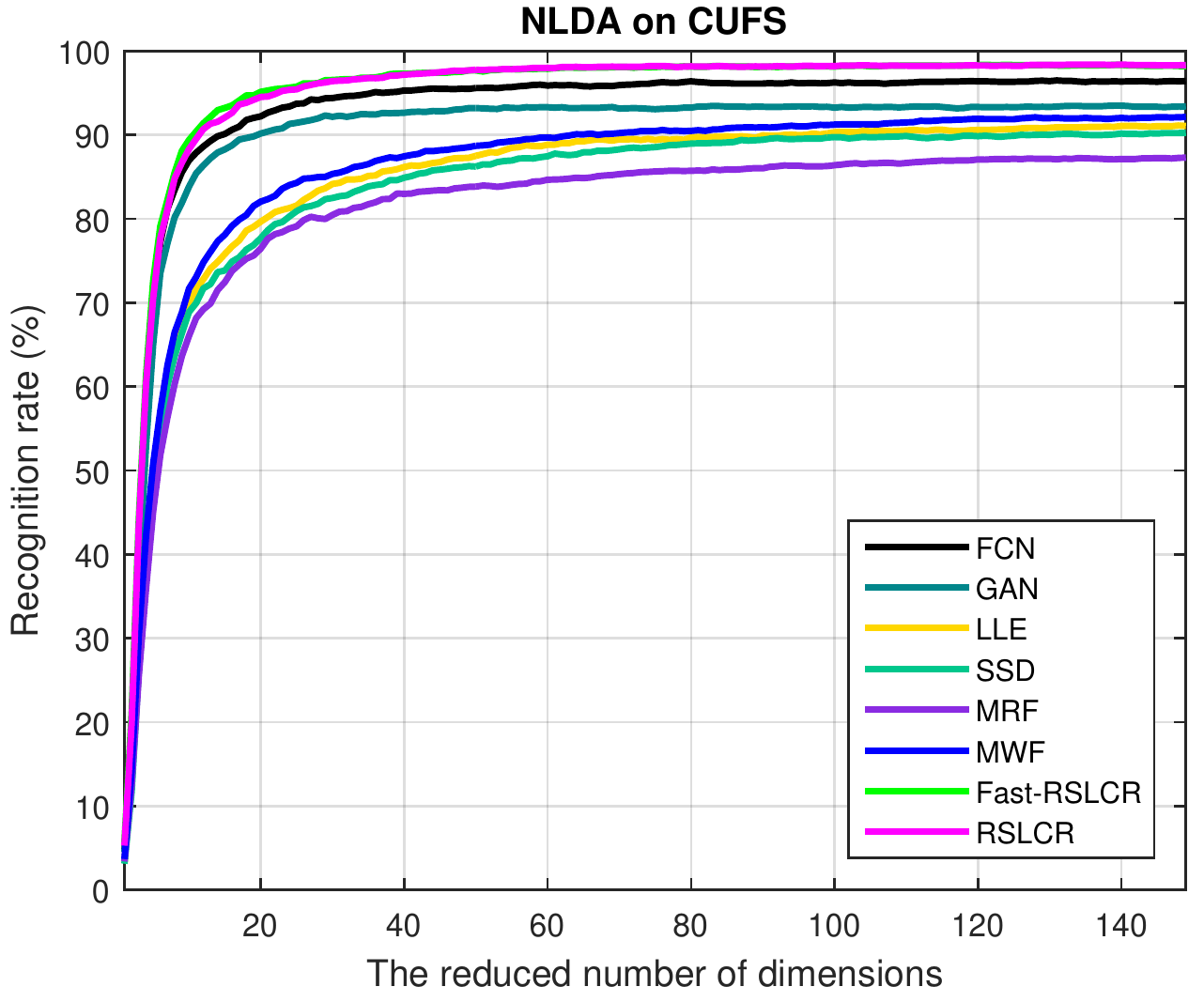}
}
\subfigure[]{
\label{fig:subfigb}
\includegraphics[width=0.45\columnwidth]{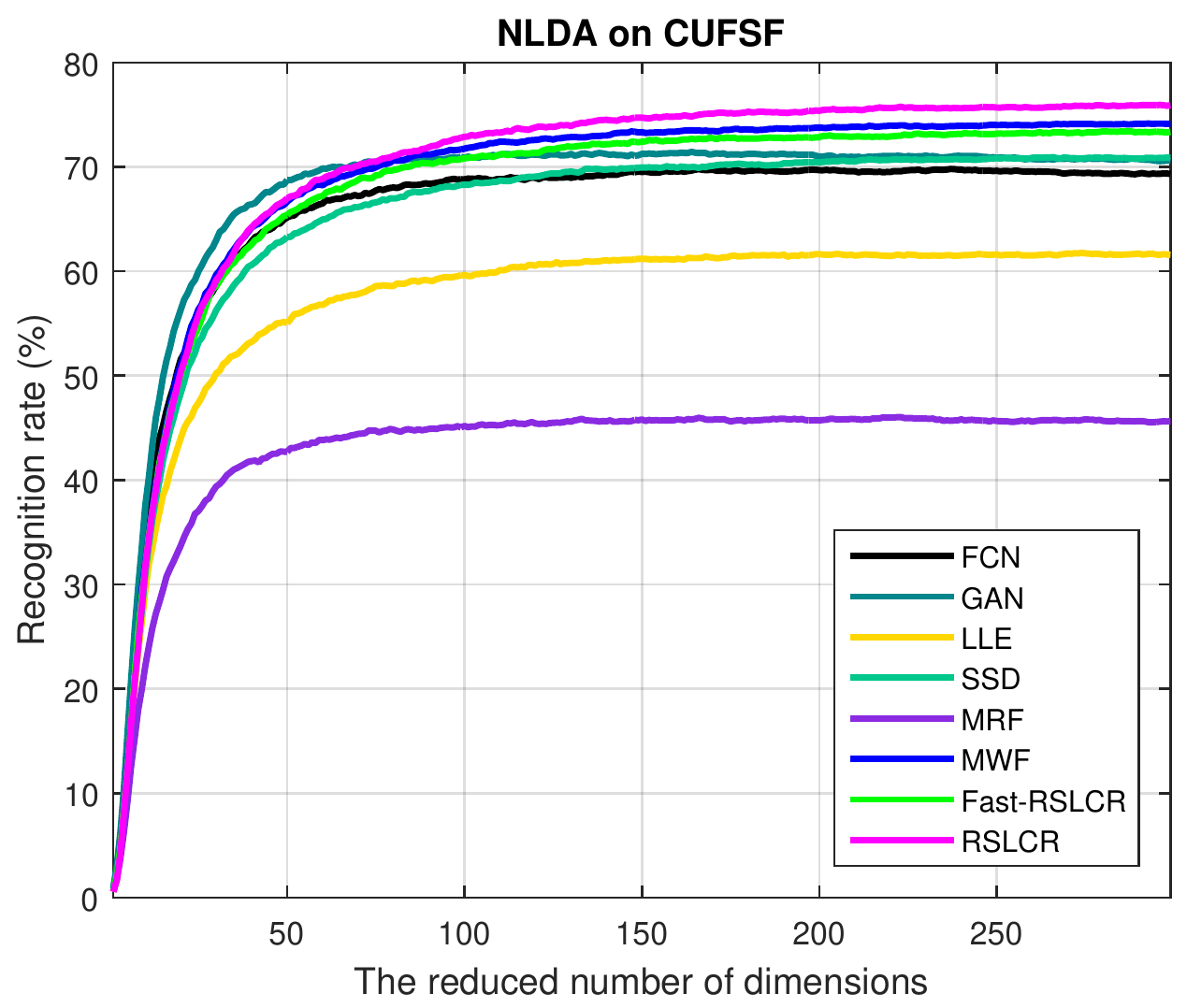}
}
\caption{ {Face recognition accuracy against variations of the number of reduced dimensions by NLDA on (a) the CUFS database and (b) the CUFSF database.}}
\label{fig:8}
\end{figure*}

Sketch based face recognition is always used to assist law enforcement. The sketch drawn by the artist is generally taken as the probe image and synthesized sketches play the role of images in the gallery. Null-space linear discriminant analysis (NLDA) \cite{Ref28} is employed to conduct the face recognition experiments. For the CUFS database, we randomly choose 150 synthesized sketches and corresponding ground-truth sketches drawn by the artist to train the classifier. The rest 188 sketches consists of the gallery. For the CUFSF database, we randomly choose 300 synthesized sketches and corresponding ground-truth sketches for training and the rest 644 synthesized sketches consist of the gallery. We repeat each face recognition experiment 20 times by randomly partition the data.

Fig. \ref{fig:8} gives the face recognition accuracy against variations of the number of reduced dimensions by NLDA on the CUFS database and CUFSF database respectively. Table \ref{tab:3} presents the best face recognition accuracy at some dimension (the number in bracket). It can be seen that on the CUFS database, the proposed two methods outperform state-of-the-art methods a lot and on the more challenging CUFSF database, our proposed RSLCR method also obtain the best performance with an accuracy of 75.94\%. The Fast-RSLCR method has comparable performance with MWF. As shown in table \ref{tab:2}, although SSD achieves higher SSIM score than MWF, it has lower face recognition accuracy than MWF. This is because though SSD could clear face sketches (much less noise than MWF) it generates face deformations (\textit{e.g.} mouth area as shown in Fig. \ref{fig:5} and Fig. \ref{fig:6}).



\section{Conclusion}
\label{sec:4}
In this paper, we presented a simple yet effective framework for face sketch synthesis based on random sampling and locality constraint. Random sampling in the offline stage could speed up the synthesis process since there is no need to search neighbors online as done in existing methods. The locality constraint could guarantee that similar sampled photo patches have similar reconstruction weights which is neglected in existing works. Through experiments including subjective (perception on the quality of synthesized sketches) and objective (image quality assessment and face recognition) evaluations illustrate the effectiveness of the proposed methods. In addition, discussion on time consumption demonstrates that the proposed Fast-RSLCR method is most efficient method in comparison to state-of-the-art methods. In the future, we would further improve the robustness of our proposed methods by incorporating more robust features. The application of RSLCR framework on related fields is another mission on the schedule.

\ifCLASSOPTIONcaptionsoff
  \newpage
\fi

\bibliographystyle{IEEEtran}
\bibliography{./RSLCR}

\end{document}